\documentclass[lettersize,journal]{IEEEtran}
\usepackage{amsmath,amsfonts}
\usepackage{algorithmic}
\usepackage{algorithm}
\usepackage{array}
\usepackage[caption=false,font=normalsize,labelfont=sf,textfont=sf]{subfig}
\usepackage{textcomp}
\usepackage{stfloats}
\usepackage{url}
\usepackage{verbatim}
\usepackage{graphicx}
\usepackage{cite}
\usepackage{multirow}

\newcolumntype{M}[1]{>{\centering\arraybackslash}m{#1}}
\newcolumntype{P}[1]{>{\raggedright\arraybackslash}m{#1}}

\usepackage{xcolor}
\begin{document}

\title{A Multimodal  Framework for Depression Detection  during Covid-19  via Harvesting Social Media: A Novel Dataset and Method}

\author{
    \IEEEauthorblockN{Ashutosh Anshul\IEEEauthorrefmark{1} \thanks{The authors have equal contribution.}, Gumpili Sai Pranav\IEEEauthorrefmark{1}, Mohammad Zia Ur Rehman\IEEEauthorrefmark{1} , Nagendra Kumar\IEEEauthorrefmark{1}}
   \\ 
    \IEEEauthorblockA{\IEEEauthorrefmark{1}Indian Institute of Technology Indore
    \\\{ashutoshanshul01, gumpilisaipranav318\}@gmail.com \{phd2101201005, nagendra\}@iiti.ac.in}
}

\maketitle
\thispagestyle{empty}

\begin{abstract}
The recent coronavirus disease (Covid-19) has become a pandemic and has affected the entire globe. During the pandemic, we have observed a spike in cases related to mental health, such as anxiety, stress, and depression. Depression significantly influences most diseases worldwide, making it difficult to detect mental health conditions in people due to unawareness and unwillingness to consult a doctor. However, nowadays, people extensively use online social media platforms to express their emotions and thoughts. Hence, social media platforms are now becoming a large data source that can be utilized for detecting depression and mental illness. However, existing approaches often overlook data sparsity in tweets and the multimodal aspects of social media. In this paper, we propose a novel multimodal framework that combines textual, user-specific, and image analysis to detect depression among social media users. To provide enough context about the user's emotional state, we propose (i) an extrinsic feature by harnessing the URLs present in tweets and (ii) extracting textual content present in images posted in tweets. We also extract five sets of features belonging to different modalities to describe a user. Additionally, we introduce a Deep Learning model, the Visual Neural Network (VNN), to generate embeddings of user-posted images, which are used to create the visual feature vector for prediction. We contribute a curated Covid-19 dataset of depressed and non-depressed users for research purposes and demonstrate the effectiveness of our model in detecting depression during the Covid-19 outbreak. Our model outperforms existing state-of-the-art methods over a benchmark dataset by 2\%-8\% and produces promising results on the Covid-19 dataset. Our analysis highlights the impact of each modality and provides valuable insights into users' mental and emotional states.

\end{abstract}

\begin{IEEEkeywords}
Depression, Covid-19, Social Media, Multimodal Analysis, Machine Learning, Deep Learning
\end{IEEEkeywords}

\section{Introduction}
\IEEEPARstart{T}{he} novel coronavirus, designated as Covid-19, emerged towards the end of 2019. This virus quickly spread across various countries, prompting the World Health Organization (WHO) to declare it a public health emergency in late January 2020. Covid-19 causes severe acute respiratory syndrome, resulting in numerous fatalities. According to the WHO, a total of 5,722,394,451 people were affected by the virus, leading to 63,904,401 deaths. Due to its rapid transmission, governments worldwide implemented lockdowns, restricting travel and confining people to their homes. These restrictions caused financial and medical crises globally, leading many individuals to experience depression. The Covid-19 Mental Disorders Collaborators reported a 27.6\% increase in cases of major depressive disorders and a 25.6\% increase in cases of anxiety disorders worldwide throughout 2020.

Depression is a leading cause of disability worldwide and a major contributor to the global burden of diseases. According to a WHO study, it is estimated that approximately 280 million people worldwide have depression, which accounts for around 3.8\% of the population. This includes 5.0\% among adults and 5.7\% among adults older than 60 years. People with depression experience various symptoms, such as loss of pleasure or interest in activities, poor concentration, and feelings of hopelessness about the future, which may also lead to suicidal thoughts. Despite being a serious illness, many individuals remain unaware or hesitant to discuss it with others. However, social media platforms such as Twitter, Facebook, and Instagram have become popular outlets for expressing emotions and mental states. In this study, our focus is on detecting mental health using social media platforms.

Detecting depression from social media platforms comes with numerous challenges. Firstly, identifying depression-oriented actions on social media can be difficult since users engage in a diverse range of actions, such as liking, commenting, and posting. However, only a small percentage of these actions exhibit depressive symptoms, such as negative self-talk, reduced social activity, and expressions of hopelessness. Secondly, the raw data obtained from social media platforms requires advanced pre-processing to derive valuable insights. Thirdly, a significant number of users employ various means, including emojis, images, and links to other posts or blogs, to express their emotions. Drawing conclusions from these types of posts presents its own set of challenges.

Multiple approaches have been proposed in the past to address the problem of depression detection by utilizing social media data~\cite{ref9,ref20,ref24}. Many of these models use surveys and questionnaires to determine the ground truth of whether a user is depressed or not. For instance, Priya \textit{et al.}~\cite{ref9} conducted a survey through Google Forms using the Depression, Anxiety, and Stress Scale (DASS-21) questionnaire. However, conducting such surveys is time-consuming and expensive. Moreover, ensuring that the collected data adequately represents a generalized population would require conducting surveys or interviews on a very large scale, which is challenging. Furthermore, the questionnaires used to create data focus on a predefined set of symptoms as depression criteria. However, behaviors related to depression evolve rapidly over time and may not be adequately covered in the questionnaires. Most of the proposed methodologies identify features from the text of social media posts, such as tweets, and build classifier models to detect depression. For example, Schwartz \textit{et al.}~\cite{ref20} extracted n-grams, topic-based features, and emotional features. However, social media posts, including tweets, are generally short in length and may not provide enough context about the user. Moreover, these methods did not consider other modalities, such as images and user profiles, which can provide valuable information about the user's mental health.

Unlike most of the existing works, our model utilizes multiple modalities, including texts and images present in tweets, as well as analyzes user profiles, to effectively detect depression. Additionally, our model harnesses URLs in tweets to extract novel extrinsic features. The intrinsic features comprise a set of visual, topic-based, emotional, depression-specific, and user-specific information extracted for a given user. To generate visual features, our model implements a CNN-based deep learning model called the Visual Neural Network (VNN). This model processes the images present in tweets to extract visual features and also extracts any textual information embedded within the images, further enhancing its performance. To validate the efficacy of our model, we evaluated its performance on the publicly available and popular Tsinghua Dataset \cite{ref24}, comparing it with other existing methods. Our model achieved an accuracy of 93.1\%, surpassing the performance of existing state-of-the-art methods. Additionally, we created a novel Covid-19 dataset comprising user profiles and tweets posted during the Covid-19 outbreak. When evaluated on this dataset, our model achieved an accuracy of 91.7\%. Furthermore, we conducted experiments to study the contribution of each feature in improving our model and examined the effects of different parameters on its performance. Additionally, we performed qualitative analysis to validate our model using a sample user, demonstrating how the proposed features contribute to enhancing the model's performance.

The key contributions of our work can be summarized as follows:
\begin{enumerate}
\item{We present a novel multimodal framework to address the problem of detecting depression among social media users by extracting external knowledge, user profiles, and historical posts.}
\item{To provide enough context about the user's emotional state, we propose two methods: (i) an extrinsic feature by harnessing the URLs present in tweets, and (ii) extracting textual content from images posted in tweets. We also extract five sets of features belonging to different modalities to describe a user. These features are used to build our classifier model, which detects depression and takes protective action to prevent its growth.}
\item{We propose a deep learning model called the Visual Neural Network (VNN) to generate embeddings of the images posted by the user. These embeddings are then used to create a visual feature vector for prediction.}
\item{We construct a well-labeled Covid-19 dataset by crawling profile information and posts from depressed and non-depressed users on Twitter during the Covid-19 outbreak. This dataset is used to evaluate the model's performance in detecting depression among users during the Covid-19 outbreak and to perform qualitative analysis.}
\item{We evaluate the model using our novel dataset as well as an additional benchmark dataset to compare its performance with other existing state-of-the-art methods. Our model outperforms the existing methods on the benchmark dataset by incorporating multimodal information present in the tweets. It also produces promising results on the Covid-19 dataset, effectively detecting signs of depression arising during the Covid-19 outbreak.}
\item{ We analyze the impact of each modality on the model's efficacy and provide important qualitative analysis based on the Covid-19 dataset.}
\end{enumerate}

The rest of the article is organized as follows. Section 2 provides a literature review of the related works done in the past. Section 3 describes the proposed model in detail. Section 4 discusses the model’s performance over the two datasets, comparison with existing methods, performance gain analysis, qualitative analysis, and parameter sensitivity analysis. Finally, Section 5 provides a brief conclusion of our work.

\section{Related Works}

In this section, we discuss some previous works related to depression detection using social media data. Firstly, we review the related works based on the social media platform selected as the data source, the methodologies adopted for data collection, and the criteria used to establish the ground truth regarding a user being depressed. The next two subsections focus on Machine Learning and Deep Learning-based methodologies, respectively. Finally, we report on works related to depression analysis and detection during the Covid-19 pandemic.

\subsection{Depression Criteria and Dataset}

In addition to the approach adopted for classifying a user as depressed or non-depressed, most of the methods also vary based on the social media platforms selected as data sources and the criteria used to establish the ground truth. Among the referred works, Twitter is the most commonly used data source~\cite{ref1,ref10,ref34}. Other major platforms that have been utilized include Reddit~\cite{ref17,ref42}, Facebook~\cite{ref20,ref21}, and Instagram~\cite{ref8}. 

The ground truth whether a user is depressed or not can be established using multiple ways. Self-reported diagnosis, where the user discloses that he/she has been diagnosed with depression, is a widely used criterion to determine the existence of depression. Jia \MakeLowercase{\textit{et al.}}~\cite{ref13}, Shen \MakeLowercase{\textit{et al.}}~\cite{ref24} and Yazdavar \MakeLowercase{\textit{et al.}}~\cite{ref15} collected stressed and depressed users by self-disclosure over Twitter. Sekulić \MakeLowercase{\textit{et al.}}~\cite{ref14} and Trifan \MakeLowercase{\textit{et al.}}~\cite{ref4} used self-reported diagnoses by Reddit users for data collection.

Dao \MakeLowercase{\textit{et al.}}~\cite{ref5} analyzed the content published over the LiveJournal\footnote{LiveJournal website - https://www.livejournal.com/} blogging website. They identified sub-categories related to depression and autism using their descriptions and collected posts from the identified subcategories. Feuston \MakeLowercase{\textit{et al.}}~\cite{ref8} collected Instagram posts indicating depression by searching for the presence of mental illness-related hashtags.

Various works rely on user surveys to determine whether the user in question is depressed or not. Guntuku \textit{et al.}\cite{ref21} created a survey on the Qualtrics platform, which included Cohen's 10-item stress scale and demographic questions. They further extracted data from the Facebook and Twitter accounts of the users who submitted the survey. Morshed \textit{et al.}\cite{ref6} and Xu \textit{et al.}~\cite{ref7} both collected data from college students. They used surveys to gather information about the mental health of the students and also tracked their activities, such as sleep duration, social interaction, and the amount of time spent on different activities.

\subsection{Machine Learning-based Methods}

Machine Learning models have been widely used in text processing tasks such as sentence classification~\cite{ref56}. Machine Learning-based depression detection involves extracting useful features from the content in a user's post and profile and building a Machine Learning classifier to detect depression using the extracted features. Choudhury \textit{et al.}\cite{ref31} extract user engagement, emotional, and egocentric features to represent a user. These features are then fed to PCA and used to build a Support Vector Machine model to predict the presence of depression. Gui \textit{et al.}~\cite{ref48} utilize GRU and VGG to extract tweet and image features respectively. Further, they employ selector models to automatically identify relevant indicator texts and images based on the text and image features. By jointly considering textual and visual information, COMMA surpasses baseline methods and enhances detection accuracy.

Priya \textit{et al.}~\cite{ref9} used data collected via Google Forms, which consisted of questions related to depression, stress, and anxiety. They built Decision Tree, Random Forest, Support Vector Machine (SVM), Naive Bayes, and K-Nearest Neighbors (KNN) models to predict the severity of depression, anxiety, and stress present in the user. Resnik \textit{et al.}\cite{ref32} proposed the use of topic modeling for linguistic analysis. They adopted Supervised Latent Dirichlet Allocation (sLDA) and trained a linear Support Vector Regression model for depression detection. Schwartz \textit{et al.}\cite{ref20} proposed a Linear Regression model to predict the degree of depression present in a user. They represented a tweet in terms of n-grams, topics present in the tweet, linguistic features extracted using LIWC categories, and the number of words in the tweet.

 Shen \textit{et al.}\cite{ref24} extract multimodal features using the tweet and the user's profile information. They implement multimodal dictionary-based learning to create a joint representation of the user and build a machine-learning model to predict whether the given user is depressed or not. Wang \textit{et al.}\cite{ref11} trained rule-based decision tables, Bayes network, and J48 trees using features extracted from microblogs to classify the blogs as depressed or non-depressed. In their later work, Wang \textit{et al.}\cite{ref16} modeled depression detection as a node classification problem.  They defined node-level features and edge-level features, based on the user's profile, sentiment analysis of blog posts, and interaction with other users.

\subsection{Deep Learning-based Methods}

Bucur \textit{et al.}~\cite{ref53} extract text and image posts from a user's social media timeline and encode them using pre-trained image and text encoders. The encoded sequence of images and texts is then passed through a cross-modal encoder and transformer encoder, incorporating positional embeddings for relative posting time. Finally, they perform classification on the mean pooled representations to determine the user's classification. Jere \textit{et al.}~\cite{ref19} propose the use of images captured from the camera to identify hand-shivering patterns in users using a Simplified-Fast-Alexnet model. They also extract features from social media posts to determine the presence of negative sentiment. Jia \textit{et al.}\cite{ref13} proposed a model to detect stress among Twitter users by using linguistic, social, visual, and user-level features. The model employs a cross auto-encoder to learn the joint representation of the tweets, followed by a CNN model and a partially-labeled factor graph for stress detection. The work also proposes a deep neural network model to detect cross-domain depression between Twitter and Webio.

Kuo \textit{et al.}~\cite{ref51} treated each user as a time-evolving graph and used four steps for depression detection: constructing heterogeneous interactive graphs for users, extracting representations of interaction snapshots with graph neural networks, modeling snapshot sequences with attention mechanism, and performing depression detection using contrastive learning. Sekulić \textit{et al.}\cite{ref14} model the problem of depression detection among users as a document classification problem and propose a deep neural network-based Hierarchical Attention Network (HAN) for classifying a user and detecting multiple mental-related illnesses, including depression, ADHD, Anxiety, PTSD, Autism, etc. Shen \textit{et al.}\cite{ref36} aimed to address the problem of cross-domain depression detection by extracting multimodal features from Webio and applying adaptive feature transformation and feature combination approaches to detect depression among Twitter users.

Wang \textit{et al.}\cite{ref39} generated word vector embeddings from Twitter posts and profiles by passing them to an XLNet, followed by a BiGRU and attention model. They concatenated the statistical features and passed the final feature set through fully connected layers for depression detection. Wang \textit{et al.}~\cite{ref52} extracted text embedding using XLNet, textual and behavioral features, and image features and proposed a FusionNet for depression detection. Yan \textit{et al.}~\cite{ref50} proposed a fusion of text and images to detect depressive tendencies. They used an improved VGG-16 model for extracting image features and BERT, BiLSTM, and Attention-based models for extracting textual features. Explainable AI (XAI) is a growing field that aims to make AI systems more understandable and transparent. It offers benefits such as increased trust, fairness, and the ability to identify and fix errors, leading to improved accuracy and performance~\cite{ref55}. Zogan \textit{et al.}~\cite{ref46} proposed an explainable deep learning model, which uses RNN to generate tweet encoding and fused them with depression-oriented features to predict whether the user is depressed.  

\subsection{Depression Detection due to Covid-19}

Recent studies have reported a strong impact of Covid-19 on people's mental health. Numerous works aim to detect depression among users due to Covid-19 or analyze the impact of Covid-19 on the spread of depression among the public~\cite{ref41,ref18,ref29,ref33}. Basheti \textit{et al.}\cite{ref41} conducted an online cross-sectional study on students studying healthcare-oriented degrees at a university in Jordan. Validation of depression was done through the Hospital Anxiety and Depression Scale (HADS) questionnaire. They observed that 33.8\% were classified to have abnormal anxiety scores, and 26.2\% were classified to have abnormal depression scores. Several factors such as smoking, lower family income, and medication were found to have a positive effect on anxiety and depression. Gupta \textit{et al.}\cite{ref18} created a Covid-19 dataset from Twitter and processed the tweets to identify topics using LDA. They also analyzed sentiment valence and emotion intensity attributes using five machine learning-based algorithms collectively named CrystalFeel. The sentiment valence intensity was converted into the sentiment category attribute, and the emotion score was converted into the emotion category attribute. 

Mathur \textit{et al.}\cite{ref22} performed an emotional analysis of Twitter posts during the Covid-19 outbreak. They explored the tweets using lexicon-based emotion detection, analyzing the emotions conveyed by a collection of words using pre-existing keywords. Ni \textit{et al.}\cite{ref30} conducted a survey using WeChat among adults and health professionals, assessing the presence of depression using the Patient Health Questionnaire-2. Their findings revealed that one-fifth of the respondents reported the presence of depression. Porter \textit{et al.}\cite{ref33} conducted a phone survey among younger individuals who grew up in poverty in low/middle-income countries (LMICs) such as India, Vietnam, and Peru. They analyzed the impact of multiple factors like food insecurity and economic adversity on anxiety levels, reporting that the mental health of young people in LMICs is affected by pandemic-related health.

Zhou \textit{et al.}\cite{ref23} proposed multimodal features and a TF-IDF-based depression detection model to analyze depression trends arising due to Covid-19 in Australia. They found a strong correlation between the spread of the pandemic and the trend of depression among the public across Australia. Zogan \textit{et al.}~\cite{ref47} introduce HCN, a hierarchical convolutional attention network for depression detection on social media. They utilized CNN+MLP for generating work encoding followed by a CNN model for tweet encoding generation. Additionally, the research investigates the influence of the COVID-19 pandemic on the identification of depression, uncovering a higher occurrence of depression during this global health crisis.

All of the aforementioned depression detection methods primarily process the text of social media posts, with a few of them also analyzing user profile information. However, social media posts are generally short in length and may often lack textual content, leading to data sparsity issues and reduced performance. Only a few studies have explored the use of multimodality, particularly extracting visual features from images posted by users. In this paper, we aim to address data sparsity and add more context to social media posts by exploring the use of external data extracted through URLs and optical character recognition. Additionally, we extract data from images posted in tweets and focus on leveraging multimodal information to enhance depression detection efficacy.

\section{Proposed Methodology}

\begin{figure}[ht]
\centering
\includegraphics[width=0.49\textwidth]{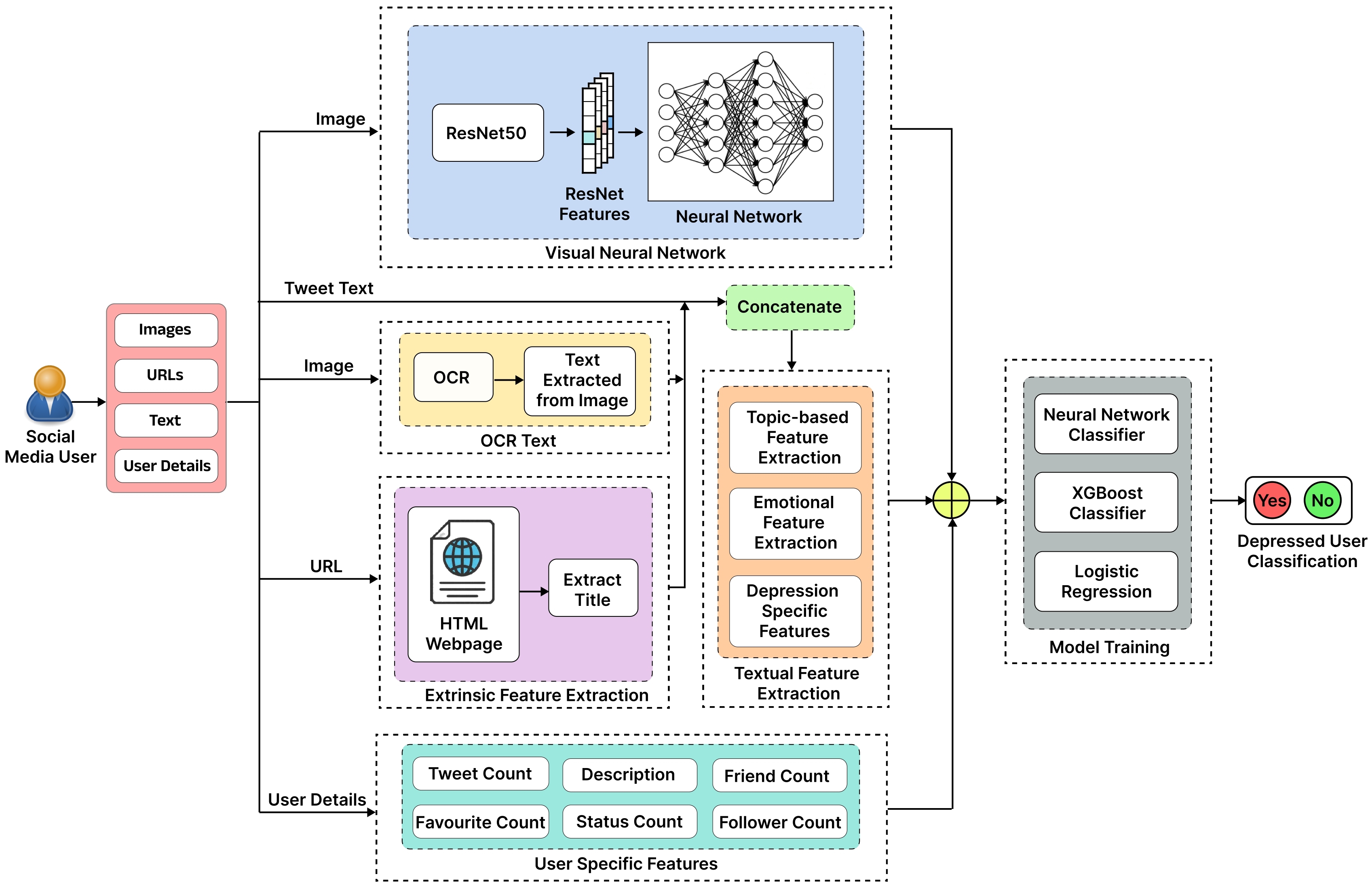}
\caption{System architecture of the proposed model}
\label{sysArch}
\end{figure}

 In this section, we present a detailed description of our Multimodal Feature-based Ensemble Learning (MFEL) method for detecting depression among users. The architecture explaining the proposed depression detection model can be seen in Fig.\ref{sysArch}. In our proposed system, we first extract user information and tweets, which are then stored in a tweet database. Next, we extract extrinsic features to gain more insights about the tweets and process them to remove any noise present. Each user is represented as a set of textual, visual, and user-specific features. These extracted features are concatenated together to create a joint representation, which is then inputted into different classifiers for depression detection. The outputs of these classifiers are further combined using an ensemble technique to achieve better performance.

\subsection{Extrinsic Feature Extraction}

Compared to normal text documents, tweets are relatively shorter in length and may lack sufficient information to analyze the user's social behavior comprehensively. Users often utilize entities such as URLs, emojis, and images in their tweets to express their thoughts and emotions, rather than relying solely on text. Extracting only textual features from such tweets may not provide enough context to effectively detect depression. Therefore, it becomes crucial to gather additional information from external sources and augment the tweet's context. Valuable information embedded within the tweets can be utilized to extract extrinsic features, which offer deeper insights into user sentiment and emotion. To extract extrinsic features, we leverage the information present in the tweets from the world wide web, allowing us to gain a more comprehensive understanding of the user's mental state.


    Users often provide online references, such as blog posts and articles, in their tweets to express their thoughts and share information. We have observed that depressed users tend to include URLs in their tweets that often lead to online resources related to depression or anxiety. These URLs can provide valuable contextual information about the user's mental health. To capture this extrinsic feature, we extracted the HTML webpage content associated with the URL provided in the tweet. By extracting the title of the webpage, which sets the precise context of the content, we appended it to the original tweet for further intrinsic feature extraction.

\subsection{Intrinsic Feature Extraction}

We extract intrinsic features from the content present in the tweet. This comprises five multimodal features extracted by processing the tweets.

\begin{enumerate}
    \item \textbf{Visual Feature:} Using images, a user can tell a story or express a thought much better than a written textual tweet. Therefore, images are an important feature for analyzing users' social behavior. In our work, we propose two methodologies to process the images associated with a user's posts:
    
    (a) \textit{Optical Character Recognition} - We have observed that a significant portion of the images posted by users contain textual content. Extracting the text embedded in these images can provide valuable context regarding users' sentiments and mental conditions. Additionally, it helps overcome the data sparsity issue when the tweet text is absent or too short to provide sufficient information about the user's emotions. To achieve this, each image is first subjected to Optical Character Recognition (OCR) processing. The OCR algorithm analyzes the connected components within the image to identify character outlines. These outlines are then grouped into blobs, which are used to identify text lines and subsequently break the text into individual words. These words are then passed through a classifier for recognition. Once the embedded text is extracted, it is appended to the tweet associated with the image. After processing all the images posted in a tweet, the resulting text is used for further textual feature extraction.
    
    (b) \textit{Visual Encoding} - In addition to textual images, users often post images that contain little or no text. However, these images can still provide valuable information for identifying whether a user is depressed or not. Depressed users tend to exhibit distinct characteristics in such images, including differences in brightness, contrast, and content, compared to non-depressed users. While previous methods, such as the one proposed by Shen \MakeLowercase{\textit{et al.}}~\cite{ref24}, focused on processing profile images or avatars, these images may not accurately represent a user's current mental state if they were set before the onset of depression and remained unchanged. In contrast, images posted along with tweets can provide contextual information and insights into the user's sentiments at the time of posting. Therefore, we propose to extract visual features from the images posted in tweets. To achieve this, we resized each image to $112\times112$ and fed it into a novel CNN-based model called Visual Neural Network (VNN). Our model builds upon the widely used image classification model, ResNet50 \cite{ref25}, using transfer learning techniques. We trained the model on a balanced subset of images posted by both depressed and non-depressed users. After training, each image is passed through the CNN model, and the output produced by the 128-dimensional dense block is extracted.
    
    \begin{figure*}[ht]
    \centering
    \includegraphics[width=5.0in]{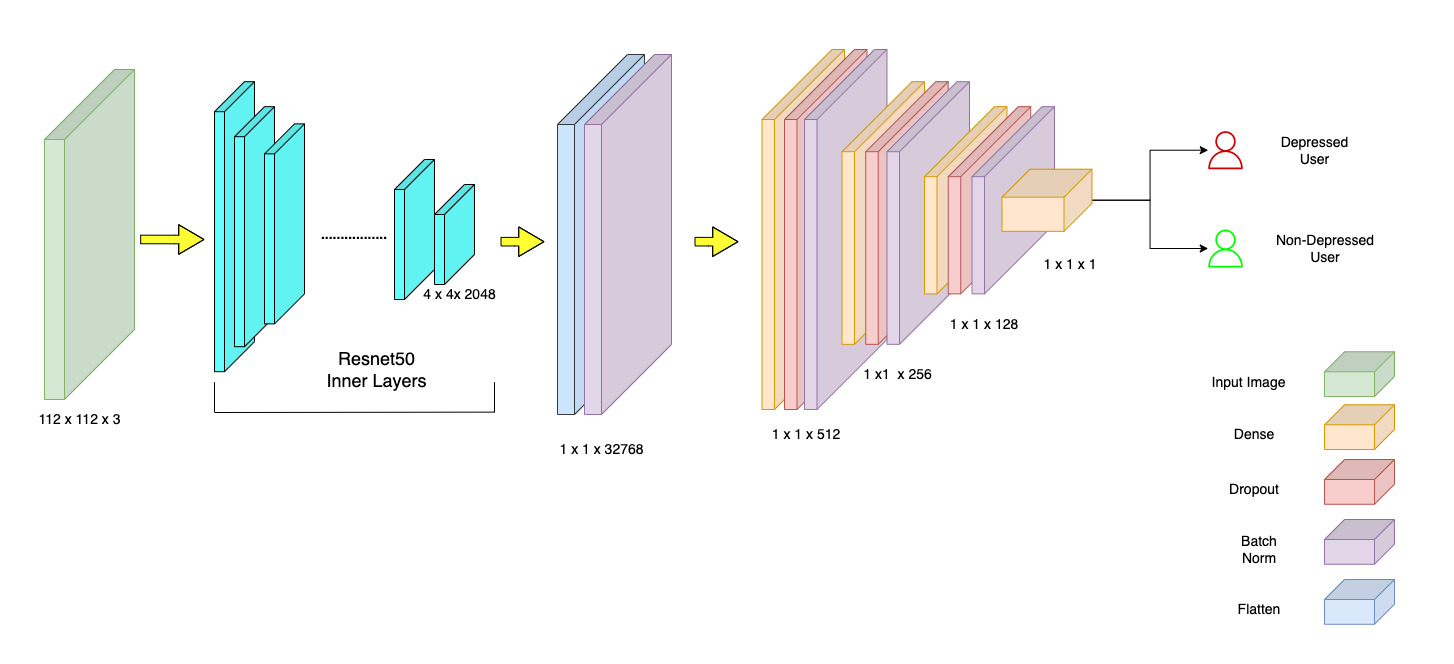}
    \caption{Architecture of Visual Neural Network for visual feature extraction}
    \label{cnnModel}
    \end{figure*}
    
   The architecture of our VNN model is depicted in Fig. \ref{cnnModel}. The model takes an input image of size $112\times112\times3$ and passes it through the ResNet layers. The output of the ResNet is a feature matrix with dimensions $4\times4\times2048$. This matrix is then flattened into a vector of size $32768$. Subsequently, the vector flows through three dense blocks, each comprising a dense layer, a dropout layer, and a batch normalization layer. The number of units in the first, second, and third dense layers is set to 512, 256, and 128, respectively. Finally, the 128-dimensional feature vector is fed into the output layer, which predicts whether the image is posted by a depressed user or not.
    
    \item \textbf{Topic-based Feature:} Topic modeling is an effective text-mining tool used to extract the hidden semantics of a document in the form of topics, which are further represented using similar words. It can be used to encode the tweets posted by an individual into a fixed set of topics and identify the presence of topics related to anxiety or depression, which would be useful in detecting depression. To achieve this, we adopted the most widely used probabilistic generative model, Latent Dirichlet Allocation (LDA) \cite{ref26},  which tries to assign topics to a document based on the assumption that every document is a mixture of topics and every topic is a mixture of terms. The probability distribution, $\phi_{td}$ of a topic $t$ provided a document $d$ can be calculated using Eq~\ref{doctopic}:
    
    \begin{equation}
        \label{doctopic}
        \phi_{td} = \frac{m_{td} + \alpha}{\sum_{i=1}^{T} m_{id} + T\alpha}
    \end{equation}
    
    where, $T$ is the number of topics, $\alpha$ is the smoothing constant, and $m_{ij}$ is the count of occurrences when topic $i$ is assigned to document $j$. Similarly, the probability distribution, $\gamma_{wt}$ of a term $w$ provided a topic $t$ can be calculated using Eq~\ref{termtopic}:
    
    \begin{equation}
        \label{termtopic}
        \gamma_{wt} = \frac{n_{wt} + \eta}{\sum_{i=1}^{W} m_{it} + W\eta}
    \end{equation}
    
    where, $W$ is the number of terms, $\eta$ is the smoothing constant, and $n_{ij}$ is the count of occurrences when term $i$ is assigned to topic $j$. We set the number of topics, $T$ to be 15. We also observed that the model identified topics better when it is trained with tweets posted by depressed users only.
    
    \item \textbf{Emotional Feature:} The emotions of a user have a significant impact on his/her mental behavior and thus impact his/her tweets. In our study, we used three different techniques to extract emotional features:
    
    (a) \label{emintense} \textit{Emotion Intensity} - Emotional intensity refers to the degree of magnitude of emotional responses. Analyzing the variations in emotional intensities can provide valuable insights into users' mental states. For instance, individuals with higher intensities of negative emotions are more likely to experience depression compared to those with lower intensities. The tweets contain information about the intensity of multiple emotions and the user's sentiments at the time of posting. We extracted the intensities of ten different emotions, including Joy, Fear, Anger, Anticipation, Disgust, Trust, Surprise, Positive, Negative, and Sadness, from the tweet content. To extract these emotional intensities, we employed a dictionary-based approach that maps lexicons to their corresponding emotions. The frequency of occurrence of each emotion can represent its intensity.
    
    To achieve this, we adopted the NRClex\footnote{NRCLex library - https://pypi.org/project/NRCLex/}\cite{ref38} that uses an affect dictionary consisting of approximately 27,000 words, and is based on the National Research Council Canada (NRC) affect lexicon and the NLTK library's WordNet synonym sets. 
    
    (b) \textit{Emoji’s Sentiment} - We observed that many users utilize emojis to convey emotional cues in their tweets, making them valuable in understanding the user's emotional state \cite{ref40}. To extract useful features from emojis, we first mapped each emoji to its corresponding sentiment score, categorizing it as positive, negative, or neutral. This mapping enabled us to convert each emoji into a three-dimensional feature representation. The overall sentiment score of emojis in a tweet is calculated by averaging the sentiment scores of each individual emoji present in the tweet. 
        
    (c) \label{lexcat} \textit{Lexicon Categories} - The words used in tweets by individuals contain valuable information regarding the presence of anxiety or depression. For instance, depressed users are more likely to use first-person singular pronouns compared to non-depressed users. To extract information about the types of words used in tweets, we employed a dictionary of categories, where each category is associated with a set of words. Our model utilizes this dictionary to identify the categories corresponding to each word and calculates the frequency of occurrence for each category in the tweet. We extracted category counts from the tweets using the dictionary-based Empath library \cite{ref37}. This library provides a 194-dimensional feature set, which we reduced to 90 dimensions using Principal Component Analysis (PCA).
    
    \item \textbf{Depression Specific Feature:} To extract features specific to depression, we compiled a set of keywords directly associated with depression and anxiety from various online sources. These keywords were utilized to create a distinct lexicon category within Empath. Additionally, Empath extends the keywords by leveraging its own Vector Space Model, which is constructed using neural embeddings generated by the skip-gram network. Furthermore, the Empath model identifies the usage of depression-related words in user tweets as a separate lexicon category through its updated dictionary.
    
    \item \textbf{User Specific Feature:} Depressed users tend to exhibit reduced activity and limited social interactions on social media compared to non-depressed users. Therefore, features related to user activity on Twitter can be effective in detecting depression. In our approach, we extracted several features related to user activity, including the number of tweets posted by the user during a one-month period, follower count, friend count, favorite count, and status count. Additionally, we considered the user description, which often provides insights into the user's emotional inclination. We extracted emotional intensity information from the user description using the previously discussed method.
    
\end{enumerate}

For each user, we extracted all the images posted by the user and used our VNN model to extract features of 128 dimensions. Since each user may have a dynamic number of images posted, we averaged out the feature vector for all the images posted by a user. In case a user posted no image during the entire timeline, we have passed a zero vector to the model. Given the tweets and user details, post feature extraction, an individual can be represented as a multimodal feature set, $X_{i} = [x_{i1}, x_{i2}, ...., x_{im}] \in R^{D}, x_{ij} \in R^{D_{j}}$, where $D_{j}$ is the dimension of $j^{th}$ modality, m is the total number of modalities, D is the dimension of the feature set $X_{i}$, and $x_{ij}$ is the feature set of $i^{th}$ user in $j^{th}$ modality. The complete feature set of all the users can be denoted by $X = [X_1, X_2, ...., X_N] \in R^{N \times D}$, where N is the total number of users.

\subsection{Depression Classification}

With the obtained feature set for all the users, we trained Logistic Regression, XGBoost, and Neural Network classifiers to detect depressed users among the given dataset. We observed that the ensemble of all three classifiers performed better than the individual classifiers. We, therefore, used an ensemble-based max voting technique to classify the mental state of a given user.

\begin{itemize}
    \item \textbf{Logistic Regression (LR):} It is a statistical model that estimates the probability of an event occurring by representing the event as a function of a set of independent variables, called features. It models the probability by mapping the linear combination of the feature vector to the sigmoid function given by Eq \ref{sigmoid}:
    
    \begin{equation}
    \label{sigmoid}
        p(x) = \frac{1}{1+\exp{(Wx+b)}}
    \end{equation}
    
    where x is the input feature vector, W is the weight vector, and b is the bias. During training, the model learns the optimal value of the weight vector W and bias b by minimizing the negative log-likelihood function given by Eq \ref{logisticloss}:
    
    \begin{equation}
        \label{logisticloss}
        L_{logistic} = \sum_{i=1}^{N} -y_{i}\log p_{i} -(1-y_{i})\log (1-p_{i})  
    \end{equation}
    where, $N$ is the number of samples in the training dataset, $y_i$ is the actual label of the $i^{th}$ sample and $p_i$ is the probability value for $i^{th}$ sample estimated using Eq \ref{sigmoid}. In our case, given the feature vector calculated for a user, the logistic regression model learns to accurately estimate the probability of the event that the given user is depressed.
    
    \item \textbf{XGBoost Classifier (XGB):} The Extreme Gradient Boosting (XGBoost) classifier is an ensemble of multiple decision tree models that uses boosting techniques to create a robust model from weaker ones.

    A decision tree model is structured like a tree, where internal nodes represent attribute tests, branches represent test outcomes, and leaf nodes indicate class labels. The decision tree learns by splitting the original dataset into subsets based on attribute tests and recursively repeating this process until no further splits yield additional value.

    Boosting is an ensemble technique that builds a series of weak models. Each model is constructed to correct the errors made by the previous models by focusing on examples that were misclassified. Gradient Boosting is a specific type of boosting algorithm where each classifier is trained to estimate the residual error of the previous classifiers.

    XGBoost is an implementation of decision trees constructed using gradient boosting. The individual decision trees are then combined to generate accurate predictions.
    
    \item \textbf{Neural Network Model:} A neural network model is composed of a large number of individual processing units, known as neurons, organized in multiple layers. A typical neural network consists of an input layer that receives the input features, one or more hidden layers that process the information from the previous layer and pass it to the next layer, and an output layer that generates the final prediction. Each neuron in a specific layer is connected to neurons in the previous and next layers. Therefore, the value of neurons in a given layer can be expressed as a function of the connected neurons from the previous layer. These connections are characterized by weights, representing the strength of the connection. The model learns to determine the optimal weight values for each connection by optimizing a loss function.

    For our problem, we built a neural network consisting of fully connected Dense, Batch Normalization, and Dropout layers. Post-feature extraction, a user is represented by a 274-dimensional feature vector. Our neural network model passes the input feature vector through three dense blocks of sizes 256, 128, and 64. A dense block consists of a dense layer, followed by a dropout layer and then a batch normalization layer. The final dense block is followed by the output layer which predicts whether the given user is depressed. The network aims to optimize the binary cross entropy loss given by the following Eq \ref{NNLoss}:
    
    \begin{equation}
        \label{NNLoss}
        L_{BCE} = \frac{1}{N}\sum_{i=1}^{N} -y_{i}\log p_{i} -(1-y_{i})\log (1-p_{i})
    \end{equation}
    where N is the total number of users, $y_i$ is the label and $p_i$ is the probability that the given user is depressed.
    
    \item \textbf{Ensemble Modeling:} Ensemble modeling is a technique that involves combining multiple base classifier models to create a stronger and more accurate model. The goal of ensemble modeling is to reduce generalization error and improve performance compared to individual base models. In our approach, we have employed the stacking technique to construct our ensemble. All the base classifiers are trained on the same original dataset, and their predictions are then combined to generate improved results.
    
    For our use case, we trained three base models, Logistic Regression ($LR$), XGBoost ($XGB$), and a Neural Network ($NN$). Further, we use max voting to combine the prediction of the three base models. According to the max voting rule, the final prediction of the ensemble model for a given user is depressed, if any two of the three base models predict the user to be depressed, else, the user is predicted to be non-depressed. Hence, the final prediction, $pred$ of the model can be represented by Eq \ref{maxvoting}:
    \begin{equation}
        \label{maxvoting}
        pred = mode(id_{LR}, id_{XGB}, id_{NN})
    \end{equation}
    where $id_{LR}$, $id_{XGB}$, and $id_{NN}$ are 1 if the respective models, LR, XGB, and NN predict the given user to be depressed.  
\end{itemize}

\section{Experimental Evaluations}
In this section, we first discuss the experimental setup and then present the experimental results to show the performance of our method.

\subsection{Experimental Setup}

This section first discusses different datasets and their pre-processing. We later present different comparison methods followed by evaluation metrics. 
\subsubsection{Dataset and Preprocessing}

To evaluate the performance of our model, we have used two datasets.

\begin{itemize}
    \item \textbf{Tsinghua Dataset:} This dataset was originally published by Shen \MakeLowercase{\textit{et al.}}~\cite{ref24}. We applied data processing techniques to the collected data, as discussed in the next subsection. After pre-processing, the resulting data consists of 2,522 depressed users and 4,533 non-depressed users. There is a total of 2,059,326 tweets, with 312,729 posted by depressed users and 1,746,597 posted by non-depressed users.
    
    \item \textbf{Novel Covid Dataset:} We created a novel Covid-19 dataset by collecting posts from Twitter using Twitter API\footnote{https://developer.twitter.com/en/docs/twitter-api/v1}. We focused mainly on depression arising during the Covid-19 outbreak and collected tweets from May 2021 to January 2022, spanning the impact of the second wave till the onset of the third wave. Furthermore, to avoid any user from being wrongly labeled, we used the collected tweets to independently create the datasets for depressed and non-depressed users. 
    \begin{itemize}
        \item Depressed user dataset - We used self-reported diagnosis to determine whether a given user is depressed or not. To accomplish this, we employed a strict check for the presence of phrases such as ``I'm diagnosed with depression," ``I was diagnosed with depression," ``I am diagnosed with depression," or ``I've been diagnosed with depression" in any tweet. If such a tweet was found, it was considered an anchor tweet, and the user who posted it was labeled as depressed. To gain a better understanding of the user's social behavior, we then extracted all the tweets posted by the user within a one-month duration around the anchor tweet. Additionally, we collected profile information about the user to analyze their social interaction and engagement, including friend count, follower count, and tweet count.

        \item Non-depressed user dataset - To build our non-depressed database, we randomly selected users and gathered their tweets within a one-month period between May 2021 and January 2022. To ensure that the randomly selected users were not experiencing depression during that one-month period, we manually searched their collected tweets and made sure that they did not exhibit any depressive symptoms such as hopelessness or negative talk. Once we identified a user who met these criteria, we labeled them as non-depressed and collected their profile information.
    \end{itemize}
    In order to assess the reliability of our dataset, we employed Krippendorff's alpha, a statistical measure commonly used to evaluate the agreement among multiple annotators in content analysis~\cite{ref54}. Krippendorff's alpha yields a coefficient ranging from 0 to 1, with higher values indicating stronger agreement. This versatile measure accommodates various data types. In our case, three annotators were provided with the entire dataset and tasked with labeling the data by thoroughly reviewing the profiles and tweets. Our calculated Krippendorff's alpha was 0.88, indicating a significant level of agreement among the annotators and demonstrating the reliability of our dataset.
    
    After building the depressed and non-depressed user datasets, we applied data pre-processing as discussed later in the subsection. We collected a total of 760 depressed users after pre-processing, with a total of 29,439 tweets. To maintain balance in the dataset, we randomly selected 760 users from our non-depressed dataset, resulting in a total of 41,763 tweets. Since we also extracted features from the URLs and images present in the tweets, we analyzed the user-level and tweet-level statistics of URLs and images in the dataset, as shown in Table \ref{datasetstatistics}.

    \begin{table}[H]
    \caption{User-level and Tweet-level dataset statistics}
    \centering
    \begin{tabular}{|l|M{2.0cm}|}
    \hline
    \textbf{Types of Tweets/Users} & \textbf{Percentage of Tweets/Users(\%)}\\
    \hline
    Tweets with no URLs & 33.17 \\
    \hline
    Tweets with at least one URL & 66.83 \\
    \hline
    Tweets with no images & 14.32 \\
    \hline
    Tweets with at least one image & 85.68 \\
    \hline
    Users who posted no URLs & 15.74 \\
    \hline
    Users who posted at least one URL & 84.26 \\
    \hline
    Users who posted no images & 20.33 \\
    \hline
    Users who posted at least one image & 79.67 \\
    \hline
    \end{tabular}
    \label{datasetstatistics}
    \end{table}
    
\end{itemize}

The variant and flexible nature of raw tweets makes it difficult to directly process them to extract relevant features and analyze them semantically. Hence, it is necessary to pre-process these raw tweets to filter out noise data and improve the quality of the text before feature extraction. We discarded non-English tweets using Compact Language Detection\footnote{https://pypi.org/project/pycld2/}. URLs, user mentions, and special characters were removed from the tweets. Additionally, common stop words such as "a," "an," "the," etc., which do not convey significant meaning in our analysis, were removed. Finally, we considered only those tweets that contain more than five words.

\subsubsection{Comparision Methods}

To evaluate the performance of the proposed model, we compared it with the following methods:

\begin{itemize}
    \item \textbf{Cooperative Multimodal Approach to Depression Detection in Twitter (GRU + VGG-Net + COMMA):} Gui \textit{et al.}~\cite{ref48} introduce COMMA, a cooperative multi-agent reinforcement learning method for depression detection in Twitter. By jointly considering textual and visual information, COMMA surpasses baseline methods and enhances detection accuracy. The findings have significant implications for advancing depression detection in social media, leveraging the cooperative nature of the method to effectively combine textual and visual cues. This approach holds promise for developing more accurate and robust methods to detect depression and improve mental health monitoring on social media platforms.
    
    \item \textbf{Depression Detection via Harvesting Social Media: A Multimodal Dictionary Learning Solution (MDL):} Shen \MakeLowercase{\textit{et al.}}~\cite{ref24} created a well-labeled dataset of depressed and non-depressed users from Twitter, discussed in the previous subsection. They extracted features belonging to different modalities such as visual features, social features, user features, topic-level features, emotional features, and domain-level features. These features were fed to a multimodal dictionary learning model to understand some common patterns among multiple modalities and build a joint sparse representation which is further used to train a binary classifier for prediction. They also analyzed the impact of each modality and carried out some case studies to understand the behavioral differences between depressed and non-depressed users.

    \item \textbf{Detecting Community Depression Dynamics Due to Covid-19 Pandemic in Australia (MMTFIDF):} Zhou \MakeLowercase{\textit{et al.}}\cite{ref23} analyze the impact of Covid-19 on depression in Australia. They process the tweets posted by users to extract multimodal features, including emotional, topic-level, and domain-specific features. These multimodal features are then combined with the term frequency-inverse document frequency (TF-IDF) to create the final feature set. A binary classifier is trained using this feature set to detect depression. The study utilizes the dataset published by Shen \MakeLowercase{\textit{et al.}}\cite{ref24}, which was discussed in the previous subsection, to evaluate the performance of their approach. Additionally, novel data from Australia is also collected to further analyze the effect of Covid-19 on depression.

    \item \textbf{DepressionNet: Learning Multi-modalities with User Post Summarization for Depression Detection on Social Media (DepressionNet):} Zogan \textit{et al.}~\cite{ref49} present DepressionNet, a deep learning framework for depression detection on social media. By learning multi-modal features from user posts and utilizing a novel attention mechanism, DepressionNet outperforms baseline methods in capturing linguistic, sentiment, and behavioral patterns associated with depression. The evaluation on a public dataset demonstrates its superiority. The findings highlight the potential of multi-modal approaches in capturing the complexity of depression-related patterns and inform the development of more accurate and robust methods for depression detection on social media.

    \item \textbf{Hierarchical Convolutional Attention Network for Depression Detection on Social Media and Its Impact During Pandemic (HCN+):} Zogan \textit{et al.}~\cite{ref47} introduce HCN, a hierarchical convolutional attention network for depression detection on social media. By leveraging the hierarchical structure of user tweets and incorporating an attention mechanism, HCN outperforms baseline methods by capturing crucial words and tweets within contextual information. The study also examines the impact of COVID-19 on depression detection, revealing an increased prevalence of depression during the pandemic. These findings have significant implications for advancing depression detection techniques, as HCN's hierarchical approach enables effective capture of complex language patterns associated with depression, leading to more accurate methods. Additionally, the analysis of the pandemic's impact underscores the importance of timely detection and intervention.
    
    \item \textbf{Explainable depression detection with multi-aspect features using a hybrid deep learning model on social media (MDHAN):} Zogan \textit{et al.}~\cite{ref46} present MDHAN, a novel deep learning model for depression detection on social media. By combining user posts with multi-modal features and employing a two-level attention mechanism, MDHAN outperforms baseline methods and offers explainable results. The findings have implications for understanding depression factors and facilitating personalized interventions for affected users. MDHAN demonstrates promise in accurately and interpretably detecting depression on social media platforms.
\end{itemize}

\subsubsection{Evaluation Metric}
We measured the performance of our method and compared the models in terms of Accuracy, Precision, Recall, and F1 Score. All these metrics can be expressed in terms of True Positive (TP), True Negative (TN), False Positive (FP), and False Negative (FN). Using these terms, we can represent our performance metrics using Eqs. \ref{accuracy}, \ref{precision}, \ref{recall}, and \ref{f1score}

\begin{equation}
    \label{accuracy}
    Accuracy = \frac{TP+TN}{TP+TN+FP+FN}
\end{equation}

\begin{equation}
    \label{precision}
    Precision = \frac{TP}{TP+FP}
\end{equation}

\begin{equation}
    \label{recall}
    Recall = \frac{TP}{TP+FN}
\end{equation}

\begin{equation}
    \label{f1score}
    F1-Score = \frac{2*Pecision*Recall}{Precision+Recall}
\end{equation}

In our use case, TP is the number of users correctly detected as depressed, TN is the number of users correctly detected as non-depressed, FP is the number of users who are originally non-depressed but predicted as depressed and FN is the number of users who are originally depressed but predicted as non-depressed.

\subsection{Experimental Results}

In this section, we present a detailed evaluation of the proposed model using two datasets. We compare the model’s result with existing methods over the Tsinghua dataset, analyzed the impact of every feature on the model’s performance, and results obtained over the novel Covid dataset.

\subsubsection{Performance comparison over Tsinghua Dataset}
To evaluate the effectiveness of our proposed model MFEL, we compare our model’s performance with existing methods. Each of the models’ performance is measured in terms of accuracy, precision, recall, and f1-score. The comparison of our model with the existing one is reported in Table \ref{tshinghuacomparision}.

\begin{table}[H]
\caption{Performance Comparison on Tsinghua Dataset}
\centering
\begin{tabular}{|l|c|c|c|c|}
\hline
\textbf{Method} & \textbf{Precision} & \textbf{Recall} & \textbf{F1-Score} & \textbf{Accuracy}\\
                & \textbf{(\%)}      & \textbf{(\%)}   & \textbf{(\%)}     & \textbf{(\%)}\\
\hline
GRU+VGG-Net+             & 90.0              &  \textbf{90.1}           &  90.0             & 90.0\\
COMMA'2023~\cite{ref48} &&&&\\
\hline
MDL'2017~\cite{ref24}             & 84.8               &  85.0           &  84.9             & 84.8\\
\hline
MMTFIDF'2021~\cite{ref23}         & 91.2               &  89.9           &  90.3             & 90.4\\
\hline
DepressionNet             & 90.3              &  77.0           &  83.1             & 83.7\\
BERT'2021~\cite{ref49} &&&&\\
\hline
DepressionNet             & \textbf{94.1}              &  73.1           &  82.3             & 83.6\\
RoBERTa'2021~\cite{ref49} &&&&\\
\hline
DepressionNet             & 88.9              &  80.8           &  84.7             & 84.7\\
XLNet'2021~\cite{ref49} &&&&\\
\hline
HCN+'2023~\cite{ref47}             & 87.1              &  86.8           &  86.9             & 86.9\\
\hline
MDHAN'2022~\cite{ref46}             & 90.2               &  89.2           &  89.3             & 89.5\\
\hline
\textbf{MFEL}                     & 91.1      &  90.0  &  \textbf{90.5}    & \textbf{93.1}\\
\hline
\end{tabular}
\label{tshinghuacomparision}
\end{table}

From the table, it is evident that MFEL outperforms the existing methods. It achieves an error reduction of 8.3\% compared to MDL in terms of accuracy. This performance improvement is attributed to MFEL's consideration of extrinsic features and the extraction of webpage headings from URLs included in a user's tweets, which adds more contextual information to the user's emotional orientation. Additionally, MFEL also analyzes the emotional intensity in users' tweets by modeling the tweets into scores for a predefined set of ten different emotions. Moreover, for visual features, MFEL considers all the images posted by the user and also extracts any textual content present in the images using OCR. In contrast, MDL only considers the brightness, color combination, cool color ratio, saturation, and clear color ratio of the user's avatar.

The improvement of 3.6\%, 6.2\%, and 8.4\% in terms of accuracy from MDHAN, HCN, and DepressionNet models respectively can be attributed to the fact that MDHAN, HCN, and DepressionNet only process tweet text and do not consider images posted by the users. Furthermore, MFEL achieves a 2.7\% improvement over MMTFIDF in terms of Accuracy. The main reason behind this improvement is that MMTFIDF only extracts textual multimodal features from user tweets, whereas MFEL also extracts visual and user-specific features. MMTFIDF analyzes emojis and slang words sentiment in the tweet and extracts topic-level and depression domain-specific features. In contrast, MFEL also extracts lexicon categories and emotional intensities from the tweets. All of these factors contribute to the context of a user's emotional state and enhance the performance of depression detection. 

\subsubsection{Model Performance over Covid-19 Dataset}
We prepared a novel dataset during the Covid-19 outbreak to evaluate our model for detecting depression spreading due to Covid-19. Table \ref{covidcomparision} reports the performance of the individual models as well as our MFEL model.

\begin{table}[H]
\caption{Performance Result on Covid-19 Dataset}
\centering
\begin{tabular}{|l|c|c|c|c|}
\hline
\textbf{Method} & \textbf{Precision} & \textbf{Recall} & \textbf{F1-Score} & \textbf{Accuracy}\\
                & \textbf{(\%)}      & \textbf{(\%)}   & \textbf{(\%)}     & \textbf{(\%)}\\
\hline
LR              & 90.9               &  86.3           &  88.5             & 88.3\\
\hline
XGB             & 91.2               &  89.2           &  90.2             & 89.7\\
\hline
NN              & 91.9               &  81.3           &  86.3             & 86.4\\
\hline
\textbf{MFEL}   & \textbf{93.3}      &  \textbf{90.6}  &  \textbf{91.9}    & \textbf{91.7}\\
\hline
\end{tabular}
\label{covidcomparision}
\end{table}

The MFEL model produces an f1-score of 91.9\% and an accuracy of 91.7\%. In terms of precision, recall, f1-score, and accuracy, it shows an improvement of 2.4\%, 4.3\%, 3.4\%, and 3.4\% respectively over Logistic regression, 2.1\%, 1.4\%, 1.7\% and 2.0\% respectively over XGBoost, and 1.4\%, 9.3\%, 5.6\% and 5.3\% respectively over the neural network model indicating that our proposed technique enhances the overall performance of the model. Also, the results show that the model is effective in detecting depression arising during the Covid-19 outbreak.

\subsubsection{Performance Comparison: Proposed Method vs. Baseline Transformers}
In this section, we compare our method with baseline transformer models such as BERT, MentalBERT, and PsychBERT. Bidirectional Encoder Representations from Transformers (BERT)~\cite{ref43} is an architecture based on multi-layered transformer encoders. With its bidirectional training approach, BERT comprehends word semantics by taking into account the contextual information around each word. This unique capability makes BERT a cutting-edge language model that excels in a wide range of natural language processing tasks.

MentalBERT~\cite{ref44} is a specialized variant of the BERT model designed specifically for mental health applications such as depression, stress, and suicidal ideation detection. It is trained on mental health datasets collected from Reddit, allowing it to gain expertise in comprehending mental health conditions, symptoms, and related concepts. PsychBERT~\cite{ref45} is another specialized BERT model that is pre-trained on a wide range of psychological and mental health texts. It incorporates data from social media and academic sources, including a substantial collection of PubMed articles. With its comprehensive understanding and analysis of psychological language, PsychBERT is a valuable tool for natural language processing in psychology, psychiatry, and mental health research.

To evaluate the comparison, we first extracted the text embeddings from the penultimate layer of the aforementioned BERT-based models, which capture rich semantic representations of the input text. Table \ref{tabbertcomparision} reports the performance produced by individual BERT-based models over the Tsinghua and Covid-19 datasets. BERT$_{tweet}$, PsychBERT$_{tweet}$, and MentalBERT$_{tweet}$ methods depict the models built using text embeddings generated by the respective transformer models, while `BERT$_{tweet}$ + Textual Features', `PsychBERT$_{tweet}$ + Textual Features', and `MentalBERT$_{tweet}$ + Textual Features' methods depict the models built using the respective text embeddings appended along with our textual features. It can be observed from the table that our proposed method, which incorporates textual features, outperforms the model built using text embeddings generated by BERT, MentalBERT, and PsychBERT. 

\begin{table*}[!t]
\caption{Performance Comparison between baseline transformers and textual features on Tsinghua and Covid-19 Dataset}
\centering
\begin{tabular}{|P{2.5cm}|M{1.0cm}|M{1.0cm}|M{1.1cm}|M{1.0cm}|M{1.0cm}|M{1.0cm}|M{1.1cm}|M{1.0cm}|}
\hline

\multirow{2}{*}{\textbf{Method}} & \multicolumn{4}{c|}{\textbf{Tsinghua Dataset}} & \multicolumn{4}{c|}{\textbf{Covid-19 Dataset}} \\
\cline{2-9}
 & \textbf{Precision (\%)} & \textbf{Recall (\%)} & \textbf{F1-Score (\%)} & \textbf{Accuracy (\%)} & \textbf{Precision (\%)} & \textbf{Recall (\%)} & \textbf{F1-Score (\%)} & \textbf{Accuracy (\%)}\\
\hline
BERT$_{tweet}$  & 69.8  &  72.5  &  71.1 & 78.3 & 93.5  &  72.7  &  81.8 & 83.0\\
\hline
MentalBERT$_{tweet}$ & 70.4  &  71.5  &  70.9 & 78.4 & 85.0  &  81.3  &  83.1 & 82.6\\
\hline
PsychBERT$_{tweet}$  & 68.0  &  75.8  &  71.7 & 78.0 & 88.4  &  77.0  &  82.3 & 82.6\\
\hline
BERT$_{tweet}$ + Textual Features  & 88.6  &  88.1  &  88.3 & 91.4 & 93.7  &  85.6  &  89.5 & 89.4\\
\hline
MentalBERT$_{tweet}$ + Textual Features  & 90.4  &  86.9  &  88.6 & 91.8 & 89.4  &  \textbf{90.6}  &  90.0 & 89.4\\
\hline
BERT$_{tweet}$ + Textual Features  & 89.2  &  87.5  &  88.4 & 91.5 & \textbf{97.5}  &  84.2  &  90.4 & 90.5\\
\hline
\textbf{MFEL}  & \textbf{91.1}      &  \textbf{90.0}  &  \textbf{90.5}    & \textbf{93.1} & 93.3      &  \textbf{90.6}  &  \textbf{91.9}    & \textbf{91.7}\\
\hline
\end{tabular}
\label{tabbertcomparision}
\end{table*}

\subsubsection{Performance Gain Analysis}
We analyzed the impact of important features in detecting depression over the Covid-19 dataset. We measured the impact of the extrinsic features on the model’s performance followed by the impact of different modality combinations.

\begin{itemize}
    \item \textbf{Extrinsic Feature:} Extrinsic features add up to the context for depression detection by exploiting resources outside the tweet body. We used the URLs embedded in tweets to fetch the headings of the websites referred to by the user in his/her tweets. To measure the impact of these headings while detecting depression, we evaluated our model by ruling out the extrinsic features and then processing the raw tweets for training our classifier model. Table \ref{pgawxtrinsic} provides the results for this impact analysis. 
    
    \begin{table}[H]
    \caption{Performance Gain Analysis Result on Extrinsic Feature}
    \centering
    \begin{tabular}{|P{1.8cm}|M{1.0cm}|M{1.0cm}|M{1.1cm}|M{1.0cm}|}
    \hline
    \textbf{Method}                                     & \textbf{Precision (\%)} & \textbf{Recall (\%)} & \textbf{F1-Score (\%)} & \textbf{Accuracy (\%)}\\
    \hline
    MFEL (without web extraction)                  & 92.5                   &  88.5               &  90.5                 & 90.2\\
    \hline
    \textbf{MFEL (with }\textbf{web extraction)}   & \textbf{93.3}          &  \textbf{90.6}      &  \textbf{91.9}    & \textbf{91.7}\\
    \hline
    \end{tabular}
    \label{pgawxtrinsic}
    \end{table}
    
    MFEL (with web extraction) depicts our actual proposed model while MFEL (without web extraction) represents the model where these extrinsic features are not considered. From Table \ref{pgawxtrinsic}, it can be observed that extracting the extrinsic feature using the URLs embedded in the tweets enhances the model’s performance by 0.8\%, 2.1\%, 1.4\%, and 1.5\% in terms of precision, recall, f1-score, accuracy respectively. This can be supported by the argument that depressed users refer to articles related to depression and extracting the heading of such web pages adds up to the context for our problem statement.
    
    \item \textbf{OCR Feature:} We analyzed the performance gain achieved by the model by extracting text embedded in the images posted by users using Optical Character Recognition (OCR). To report the performance gain, we first built our model without feeding it the OCR output and evaluated its performance in terms of precision, recall, f1-score, and accuracy. We then compared the results with that achieved by the originally proposed model.
    
    \begin{table}[H]
    \caption{Performance Gain Analysis Result on OCR Feature}
    \centering
    \begin{tabular}{|P{1.8cm}|M{1.0cm}|M{1.0cm}|M{1.1cm}|M{1.0cm}|}
    \hline
    \textbf{Method}                             & \textbf{Precision (\%)} & \textbf{Recall (\%)} & \textbf{F1-Score (\%)} & \textbf{Accuracy (\%)}\\
    \hline
    MFEL (without OCR feature)                  & 89.6                   &  92.8               &  91.2                 & 90.5\\
    \hline
    \textbf{MFEL (with OCR feature)}    & \textbf{93.3}         &  \textbf{90.6}      &  \textbf{91.9}       & \textbf{91.7}\\
    \hline
    \end{tabular}
    \label{pgaocr}
    \end{table}
    
   Table \ref{pgaocr} presents the results of the performance gain analysis for the OCR features. MFEL (with OCR feature) represents our original proposed model, while MFEL (without OCR feature) represents the model built by skipping the OCR text extraction step. The table demonstrates that the proposed model outperforms the model without OCR feature by 3.7\% in terms of precision, 0.7\% in terms of F1-score, and 1.2\% in terms of accuracy. This improvement can be attributed to the fact that some of the images posted by depressed users contain textual information relevant to mental health and depression. Extracting the text from such images adds more context about the user's mental health and proves to be useful for depression detection. 
    
    \item \textbf{Modality Combinations:} We present an analysis of the impact of individual modalities on the model’s performance. To accomplish this, we evaluated our proposed model’s performance by feeding it different combinations of modalities for training the classifier models. These combinations were built by a leave-one-out methodology where we left one of the multimodal features and fed the modal with the rest of them. The modalities identified for feature extraction as discussed in the proposed method are (i) Visual (v), (ii) Topic-based (t), (iii) Emotional (e), (iv) Depression Specific (d), (iv) User Specific (u).
    
    \begin{table*}[!t]
    \caption{Performance Gain Analysis Result on Modality Combinations}
    \centering
    \begin{tabular}{|l|c|c|c|c|}
    \hline
    \textbf{Method} & \textbf{Precision(\%)} & \textbf{Recall(\%)} & \textbf{F1-Score(\%)} & \textbf{Accuracy(\%)}\\
    \hline
    MFEL (t+e+d+u)   & 88.5                   &  77.7               &  82.7                 & 83.0\\
    \hline
    MFEL (v+t+d+u)   & 89.8                   &  76.3               &  82.5                 & 83.0\\
    \hline
    MFEL (v+t+e+u)   & 92.1                   &  84.2               &  88.0                 & 87.9\\
    \hline
    MFEL (v+t+e+d)   & 91.0                   &  87.1               &  89.0                 & 88.6\\
    \hline
    MFEL (v+e+d+u)   & 90.4                   &  88.5               &  89.4                 & 89.0\\
    \hline
    \textbf{MFEL (v+t+e+d+u)}   & \textbf{93.3}      &  \textbf{90.6}  &  \textbf{91.9}    & \textbf{91.7}\\
    \hline
    \end{tabular}
    \label{pgamodality}
    \end{table*}
    
    Table \ref{pgamodality} presents the results of the proposed model. MFEL(v+t+e+d+u) represents our actual MFEL model. To assess the impact of the visual feature, we removed the visual feature and used only the topic-level, emotional, depression-specific, and user-specific features, represented by MFEL(t+e+d+u). Similarly, we evaluated the model with other possible modality combinations as mentioned in the table.
    It can be observed that MFEL outperforms the other models, indicating the contribution of each modality in detecting depression. The incorporation of multimodal features extracted from the user's profile and historical tweets enhances the overall performance of the proposed model. Notably, MFEL demonstrates the highest improvement compared to MFEL(t+e+d+u) and MFEL(v+t+d+u). Specifically, MFEL improves precision, recall, F1-score, and accuracy by 4.8\%, 12.9\%, 9.2\%, and 8.7\% respectively compared to MFEL(t+e+d+u), and by 3.5\%, 14.3\%, 9.4\%, and 8.7\% respectively compared to MFEL(v+t+d+u). This suggests that images posted by users contain highly discriminative information, contributing significantly to distinguishing between depressed and non-depressed individuals. Furthermore, it can be inferred that the emotional state of depressed users differs from that of non-depressed users, making the emotional modality a highly impactful feature.
\end{itemize}

\subsubsection{Qualitative Analysis}
This section provides an analysis of sample tweets posted by depressed users, exploring the modalities present in the tweets, including images, text, and external entities such as URLs. It assesses how each modality relates to the presence of depression in the user.

We examined the URLs mentioned by depressed users in their posts and collected the titles of the web pages they linked to. Here are the titles of some sample web pages tagged by depressed users:

\begin{itemize}
    \item SeaWorld Of Hurt: Where Happiness Tanks
    \item Self-harm alternatives - Stay strong
    \item National charity helping people with Anxiety - Anxiety UK
    \item Four Ways to Improve Student Mental-Health Support (Opinion)
    \item Mental Health Awareness, And Breaking Through... - Helping Artists Create and Share
    \item Madeleine Kuderick’s top 10 books that explore mental health issues | Children's books | The Guardian
    \item Celebrating My Birthday with Depression | HealthyPlace
\end{itemize}

The above titles demonstrate a strong relationship with mental health. Therefore, leveraging these URLs proves to be valuable in gaining a deeper understanding of the online resources that users typically refer to, providing more insight into their mental state. When these titles were analyzed using the lexicon categories extraction module of our model, the category "depression terms" appeared among the top three categories for most of them. This indicates that the MFEL model effectively utilizes the external sources mentioned in tweets, processing them to gather information about the user's mental state, thereby enhancing the model's performance.

\begin{figure}[!t]
    \includegraphics[width=0.49\textwidth]{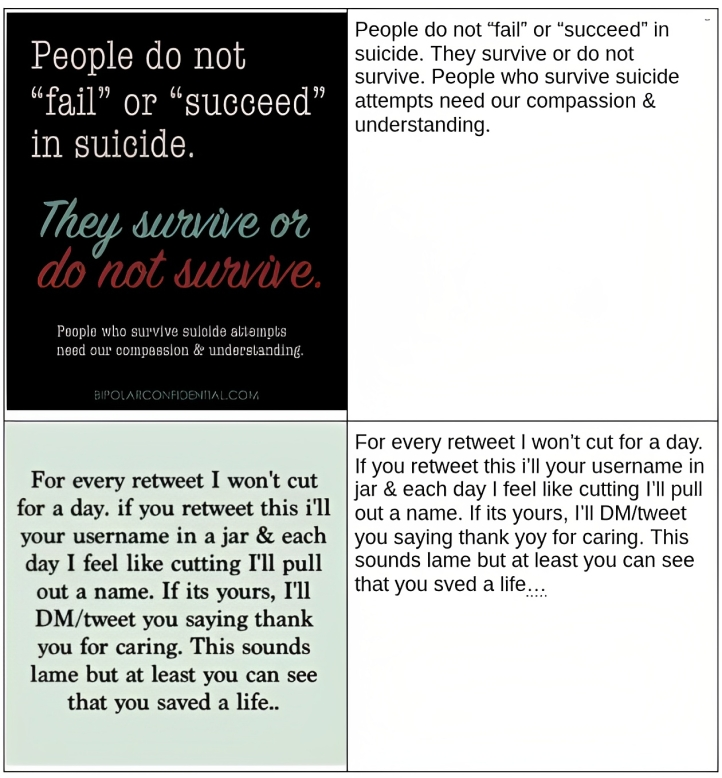}
    \caption{Sample images posted by depressed users containing textual content}
    \label{sampImages}
\end{figure}

Fig \ref{sampImages} presents images with textual content posted by depressed users. It can be seen that the text present in these images is related to suicide ideation and have a clear indication of mental illness. Extracting the textual content present in the image posted by any depressed user and appending them to the original tweet would certainly add context to the depression present in the user.

\begin{table*}[!t]
\caption{Sample tweets analysis result of a depressed user}
\centering
\begin{tabular}{|M{2.7cm}|M{6.3cm}|M{2.7cm}|}
\hline
\textbf{Tweet} & \textbf{Emotional Intensity Graph} & \textbf{Top 10 lexicon categories}\\
\hline
18 f***ing months into this pandemic and covid depression is still hitting like a truck & \includegraphics[width=0.3\textwidth]{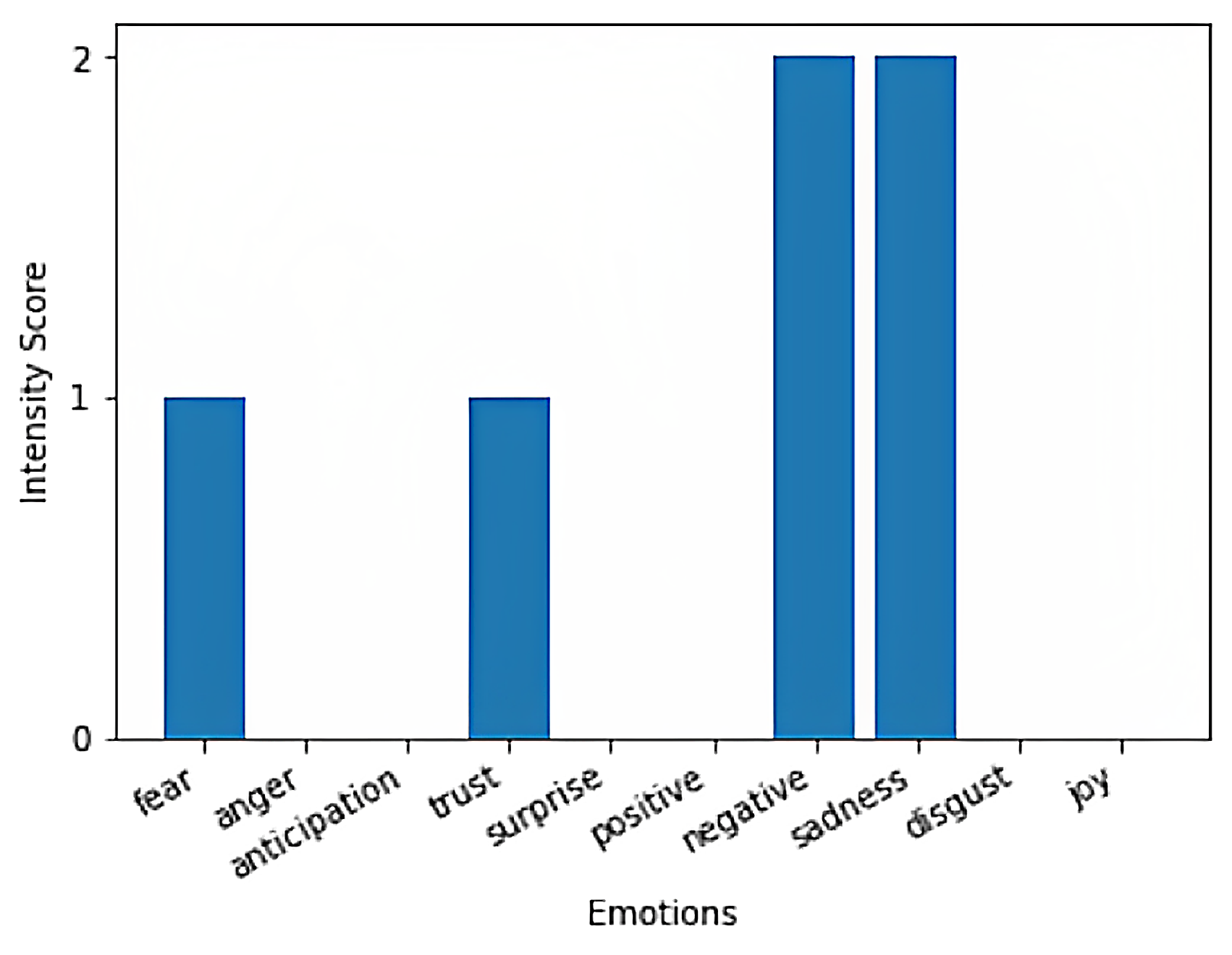} & depression\_terms, neglect, suffering, sadness, health, negative\_emotion, car, vehicle, driving, pet\\
\hline
can't go to concerts, can't see my idol in person, can't go to conferences, can't travel without f***ing quarantine here, can't even be with family when my grandma passes & \includegraphics[width=0.3\textwidth]{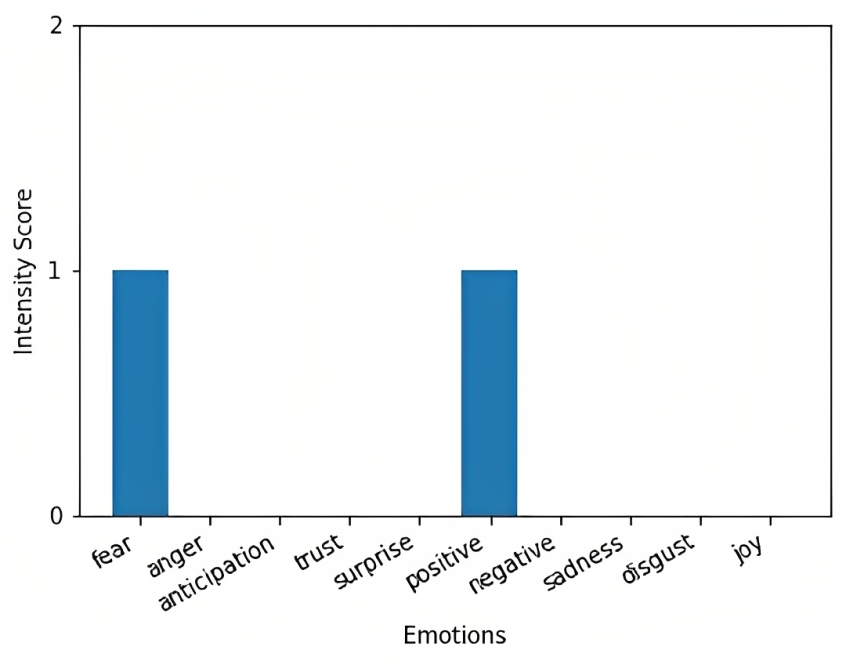} & air\_travel, traveling, movement, terrorism, tourism, negative\_emotion, vacation, surprise, hipster, religion \\
\hline
\textbf{User's Overall Result} & \includegraphics[width=0.3\textwidth]{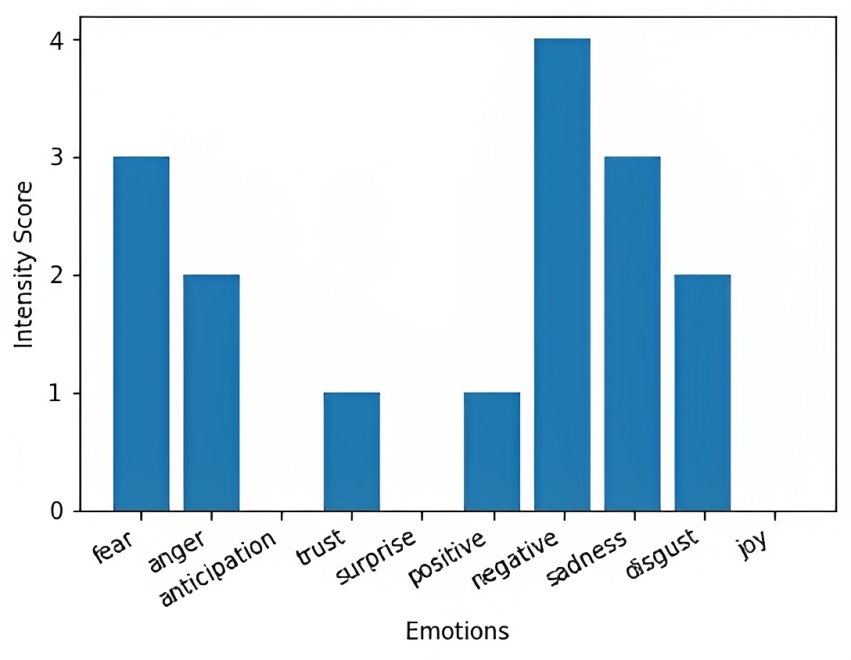} & negative\_emotion, suffering, swearing\_terms, traveling, pain, tourism, car, shame, emotional, sadness \\
\hline
\end{tabular}
\label{qualitativeuser}
\end{table*}

Table \ref{qualitativeuser} presents sample tweets of a depressed user and emotional analysis. The Emotional Intensity Graph plots the intensity scores for different emotions extracted using the dictionary-based methodology discussed in Section \ref{emintense}, under the topic Emotion Intensity. For each of the tweets, it can be seen that the score related to negative emotions surpasses that of positive ones. Furthermore, the overall graph for the user shows a distinctive gap between the score associated with negative and positive emotions. This shows that the model can effectively identify these emotional inclinations from input tweets. 

Similarly, the third column of the table reports the top 10 lexicon categories to which the user's choice of words can be categorized for each of the sample tweets. These categories are extracted using the process discussed in Section \ref{lexcat}, under the topic "Lexicon Categories". The results reveal that the top 10 identified categories include those related to negative emotions. Moreover, the overall results show that six categories, namely negative emotion, suffering, pain, shame, emotional, and sadness, are directly associated with emotional or mental states. Among them, five categories, namely negative emotion, suffering, pain, shame, and sadness, indicate a negative emotional state. This demonstrates that a user's emotional and mental state can be reflected in the words they use in their posts, and the model is able to accurately identify the presence of mental illness in the user.

\begin{table*}[!t]
\caption{Top 10 words related to each topic learned by LDA}
\centering
\begin{tabular}{|M{2.3cm}|M{2.3cm}|M{2.3cm}|M{2.3cm}|M{2.3cm}|}
\hline
\textbf{Topic 1} & \textbf{Topic 2} & \textbf{Topic 3} & \textbf{Topic 4} & \textbf{Topic 5}\\
\hline
black, didnt, month, live, need, hes, face, men, history, thread & love, day, thank, wanna, left, shit, sun, today, happy, fun & a*s, job, got, wrong, b*t*h, bad, bro, man, dude, music & dont, know, years, want, let, care, people, believe, think, feel & covid, ive, depression, gonna, days, got, getting, vaccinated, covid19, anxiety\\
\hline
\hline
\textbf{Topic 6} & \textbf{Topic 7} & \textbf{Topic 8} & \textbf{Topic 9} & \textbf{Topic 10}\\
\hline
f**k, saying, home, stay, pay, people, stuff, public, money, fine & work, time, friends, great, things, long, doesnt, people, want, took & year, old, covid, heart, school, masks, got, best, time, friend & better, trying, stop, twitter, think, sleep, red, new, thing, people & health, mental, people, problem, need, mask, wear, room, amp, lockdown\\
\hline
\hline
\textbf{Topic 11} & \textbf{Topic 12} & \textbf{Topic 13} & \textbf{Topic 14} & \textbf{Topic 15}\\
\hline
f***ing, mean, wait, read, hate, cool, arent, reading, nice, actually & new, watch, game, video, season, favorite, time, birthday, play, happy & students, times, sir, feeling, exams, private, people, worst, hope, help & good, youre, sorry, sure, pretty, going, wish, hope, new, people & people, life, wont, girl, said, instead, world, dog, hand, weird\\
\hline
\end{tabular}
\label{ldatopics}
\end{table*}

Table \ref{ldatopics} presents a list of the top 10 words that represent each of the 15 topics learned by the LDA model. Topic 10, represented by words such as `health', `mental', and `problems', and Topic 8, represented by `school' and `exam', are also useful in detecting depression as these words are associated with issues that impact mental health. Furthermore, words such as `love', `fun', `thank', and `happy' in Topic 2, and `hope', `sorry', and `good' in Topic 14 may not directly indicate depression, but they provide context for the emotional state of the user, and thus can be used to compare the sentiments of depressed and non-depressed users.

\subsubsection{COVID-related Depression vs Regular Depression}
COVID-related depression differs from regular depression in several ways. It arises from specific circumstances brought about by the pandemic, which include, social isolation, economic challenges, loss of loved ones, and the overall uncertainty surrounding the ongoing crisis. It affects a larger portion of the population and has a collective impact on mental health. The duration of COVID-related depression is often tied to the duration of the pandemic. While there are similarities with regular depression, understanding these differences can help in addressing the specific challenges and providing appropriate support to individuals experiencing COVID-related depression.

While for regular depression detection, the multimodal information present in the data would be enough, for detecting COVID-related depression, the model should also be aware of the COVID-specific context in the tweets. For instance, from Table \ref{ldatopics}, it is notable that Topic 5 is particularly relevant to COVID-19 and depression, as it includes words such as `COVID', `depression', `vaccination', and `anxiety'. Similarly, Topic 10 includes words such as `health', `mental', `mask', and `lockdown'. This suggests that users are discussing mental health issues related to the pandemic, providing valuable insights into the relationship between the pandemic and depression. Moreover, Topic 6, which includes words like `stay' and `home', indirectly indicates a lockdown scenario that can work as a stressor related to the pandemic and can contribute to the development or worsening of depression. Hence, it is important for the model to identify the presence of such context in the tweet and comprehend the relationship between the pandemic and depression.

\subsubsection{Parameter Sensitivity Study}
The optimal parameters for each of the models used were found using grid search. For the Logistic Regression model, there are four parameters: solver, regularizer ($penalty$), tolerance ($tol$), and regularization strength ($C$). The optimal values for these parameters were found using the grid search technique and the values which gave the highest accuracy were used in the final results. We observed that the Logistic Regression produced the highest accuracy when the $solver$ is `liblinear', the $penalty$ is `l2' norm, $tol$ = 0.0001, and $C$ = 10. For other classifier models, we adopted the same procedure to obtain the optimal parameter values. For the XGBoost model, we obtained the optimal performance with $max\_depth$ = 8, $learning\_rate$ = 0.2, $gamma$ = 0, $scale\_pos\_weight$ = 5, $reg\_lambda$ = 0. We performed a grid search for $batch\_size$, $learning\_rate$, and $dropout$ parameters in the neural network-based model, and they were obtained to be 32, 0.01, and 0.5 respectively with the `Adam' $optimizer$.

The three other key parameters in the model are (i) the number of topics in the Topic Based Feature ($num\_topics$), (ii) the number of components in PCA used over Lexicon Categories ($n\_components$), and (iii) the length of the visual feature obtained from our CNN model ($image\_shape$). The optimal value for each of them was found by analyzing the model’s performance on varying one of the features while keeping the other two constants.

\begin{figure}[!t]
    \includegraphics[width=0.49\textwidth]{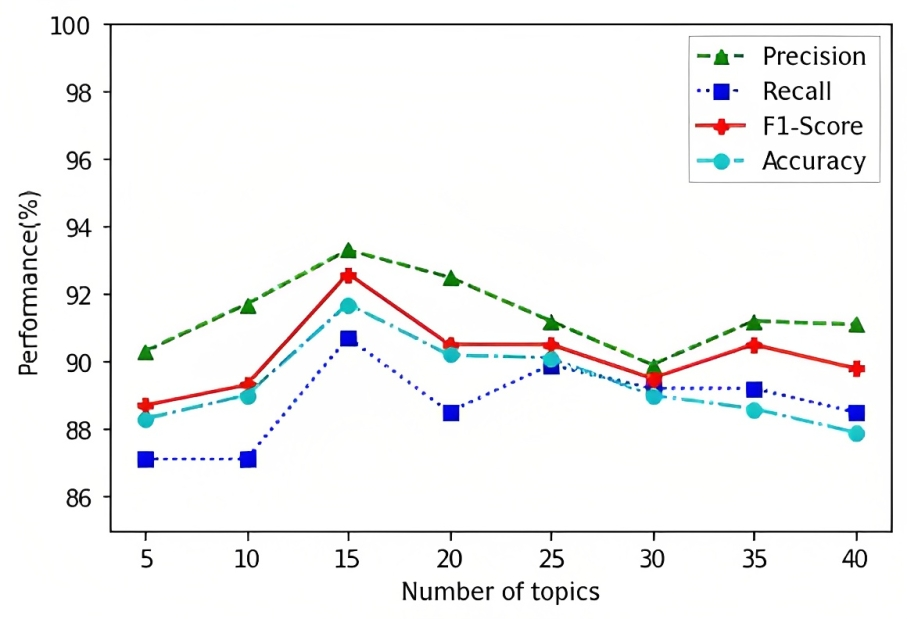}
    \caption{Effect of number of topics in Topic-Based features over model performance}
    \label{ldaGraph}
\end{figure}

    Fig \ref{ldaGraph} shows the effect of the model's performance by changing the number of topics to be extracted in the Topic-based feature. We varied the number of topics from 5 to 40. It can be observed from the graph that the performance increases first, with the increasing number of topics, and reaches its maximum when the number of topics reaches 15. The performance reduces when the number of topics is increased further. Hence the optimal number of topics into which the tweet text is to be grouped is selected to be 15.

\begin{figure}[!t]
    \includegraphics[width=0.49\textwidth]{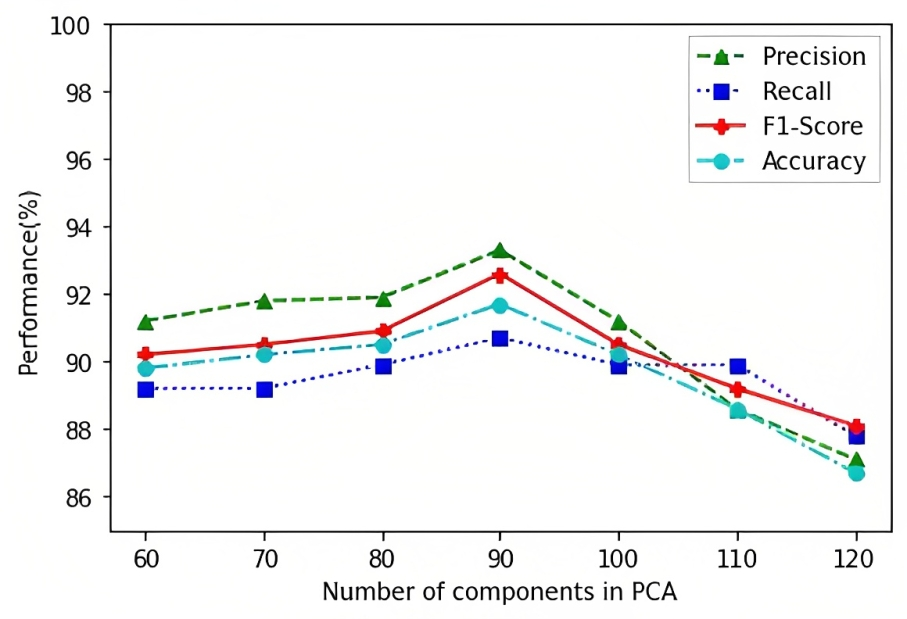}
    \caption{Effect of number of components in PCA for reducing the dimension of Lexicon categories}
    \label{pcaGraph}
\end{figure}

A similar trend was observed when we evaluated our model by varying the number of components in our PCA model. The performance analysis is shown in Fig \ref{pcaGraph} The model's performance increased and reached the maxima when the number of components reached 90 and then went downhill. Hence, the optimum number of components for our PCA model is selected to be 90.

\begin{figure}[!t]
    \includegraphics[width=0.49\textwidth]{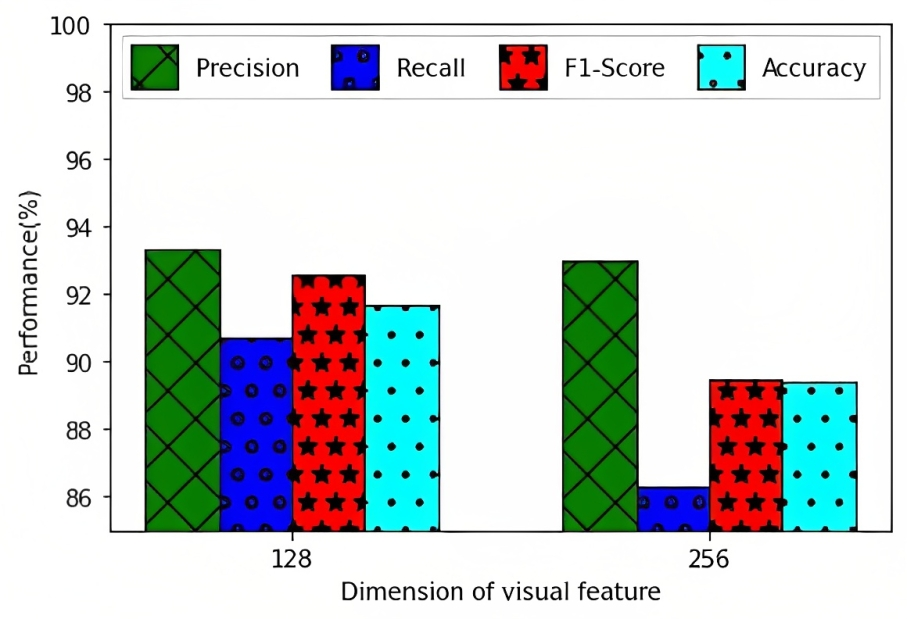}
    \caption{Effect of the dimension of Visual features over model performance}
    \label{imshapeGraph}
\end{figure}

We also examined the impact of varying the dimensionality of the visual features on the model's performance. Given that the last two dense blocks of our CNN model had sizes of 256 and 128, we evaluated the model using both 256-dimensional and 128-dimensional visual features. Figure \ref{imshapeGraph} illustrates the graphical comparison between the two options. It is evident that the 128-dimensional visual feature outperforms the 256-dimensional feature in terms of precision, f1-score, and accuracy.

\section{Conclusion}
In this paper, we propose a multimodal learning technique that combines intrinsic and extrinsic features for robust depression detection. Our approach incorporates visual and extrinsic features from tweets, as well as topic-based, emotional, and depression-specific features from textual content. We also extract user-specific features to capture individual characteristics. By employing an ensemble approach that combines Deep Learning and Machine Learning techniques, we achieve promising detection results. Our contributions include the introduction of extrinsic features derived from URLs and textual content in images posted in tweets to address the data sparsity issues in existing methods, as well as the use of a Visual Neural Network (VNN) for generating visual feature vectors. We curate a well-labeled Covid-19 dataset to evaluate our model's performance in detecting depression during the pandemic. Experimental results on real-world social media datasets demonstrate the superiority of our approach compared to existing methods in terms of accuracy and demonstrate the contribution of each modality toward enhancing the model's performance. Overall, our study highlights the importance of considering multiple modalities and external data sources in depression detection, showcasing the effectiveness of our proposed model. By leveraging a combination of textual, visual, and user-specific features, we can gain deeper insights into users' mental health and contribute to early detection and intervention efforts.

\section*{Data Availability and Access}
\label{sec:data}
We have used two datasets for our experiments. The first dataset is published by Shen \MakeLowercase{\textit{et al.}}~\cite{ref24}. Our second novel Covid-19 dataset is currently not publicly available. However, after our paper's publication, we plan to upload the same on our GitHub repository\footnote{https://github.com/AshutoshAnshul/Depression-Detection}.

\section*{Declaration}
\label{sec:coi}
Conflict of Interests: The authors have no conflict of interests that are relevant to the content of this article.

\newpage

\vfill

\end{document}